\documentclass[10pt,twocolumn,letterpaper]{article}

\usepackage{cvpr}              %

\usepackage{graphicx}
\usepackage{amsmath}
\usepackage{amssymb}
\usepackage{booktabs}
\usepackage{bbm}

\usepackage[accsupp]{axessibility}  %

\usepackage[pagebackref,breaklinks,colorlinks]{hyperref}

\def\cvprPaperID{10294} %
\def\confName{CVPR}
\def\confYear{2023}

\usepackage{amsmath}
\usepackage{mathtools}
\usepackage{overpic}
\usepackage{color}
\usepackage{listings} %
\usepackage[skip=.2em]{caption} %
\usepackage{xcolor}
\usepackage{currfile}
\usepackage{cancel}
\usepackage{float}
\usepackage{rotating}

\usepackage[rightcaption]{sidecap}  %
\sidecaptionvpos{figure}{t} %
\definecolor{turquoise}{cmyk}{0.65,0,0.1,0.3}
\definecolor{purple}{rgb}{0.65,0,0.65}
\definecolor{dark_green}{rgb}{0, 0.5, 0}
\definecolor{orange}{rgb}{0.8, 0.6, 0.2}
\definecolor{red}{rgb}{0.8, 0.2, 0.2}
\definecolor{darkgray}{rgb}{0.5, 0.5, 0.5}
\definecolor{darkred}{rgb}{0.6, 0.1, 0.05}
\definecolor{blueish}{rgb}{0.0, 0.3, .6}
\definecolor{light_gray}{rgb}{0.7, 0.7, .7}
\definecolor{pink}{rgb}{1, 0, 1}
\definecolor{greyblue}{rgb}{0.25, 0.25, 1}

\newcommand{\At}[1]{\marginpar{\tiny{\textcolor{blueish}{ANDREA}}}}
\newcommand{\Df}[1]{\marginpar{\tiny{\textcolor{greyblue}{DAVID}}}}

\newcommand{\eq}[1]{(\ref{eq:#1})}

\usepackage{bbm}
\usepackage{lipsum}
\usepackage{blindtext}

\renewcommand{\paragraph}[1]{\vspace{.5em}\noindent\textbf{#1}.}

\newcommand{\NeRF}{NeRF\xspace}
\newcommand{\RobustNeRF}{RobustNeRF\xspace}
\newcommand{\mipNeRFthreesixty}{mip-NeRF 360\xspace}
\newcommand{\ddnerf}{D$^2$NeRF\xspace}
\newcommand{\DDNeRF}{\ddnerf}
\newcommand{\NeRFW}{NeRF-W\xspace}
\newcommand{\SupplementaryMaterial}[1]{{supplementary material}}

\usepackage{enumitem}
\setlist[itemize]{noitemsep,leftmargin=*,topsep=0em}
\setlist[enumerate]{noitemsep,leftmargin=*,topsep=0em}

\newcommand{\loss}[1]{\mathcal{L}_\text{#1}}
\newcommand{\expect}{\mathbb{E}}

\newcommand{\C}{\mathbf{C}} %
\newcommand{\oracle}{\mathbf{S}}
\newcommand{\segmenter}{\mathcal{S}}
\newcommand{\image}{\C}

\newcommand{\params}{\boldsymbol{\theta}}

\newcommand{\radiance}{\mathbf{c}}

\newcommand{\ray}{\mathbf{r}}
\newcommand{\pose}{\mathbf{T}}
\newcommand{\origin}{\mathbf{o}}
\newcommand{\dir}{\mathbf{d}}
\newcommand{\position}{\mathbf{x}}

\newcommand{\density}{\sigma}

\newcommand{\kernel}{\kappa}
\newcommand{\weight}{\omega}

\newcommand{\residuals}{\boldsymbol{\epsilon}}
\newcommand{\residual}{\epsilon}

\newcommand{\threshold}{\mathcal{T}}

\begin{document}

\newcommand{\thetitle}{RobustNeRF: Ignoring Distractors with Robust Losses$^4$
}
\title{\thetitle}

\author{
Sara Sabour$^{1,2}$ \quad
Suhani Vora$^{1}$ \quad
Daniel Duckworth$^{1}$ \quad
Ivan Krasin$^{1}$ \\
David J.\ Fleet$^{1,2}$ \quad
Andrea Tagliasacchi$^{1,2,3}$
\\
\small{
\textsuperscript{1}Google Research, Brain Team
\quad
\textsuperscript{2}University of Toronto
\quad
\textsuperscript{3}Simon Fraser University
}
}

\maketitle
\footnotetext[4]{Work done at Google Research.}
\begin{abstract}
\vspace*{-0.25cm}
Neural radiance fields (\NeRF) excel at synthesizing new views given multi-view, calibrated images of a static scene.
When scenes include distractors, which are not persistent during image capture (moving objects, lighting variations, shadows), artifacts appear as  view-dependent effects or 'floaters'.
To cope with distractors, we advocate a form of robust estimation for NeRF training, modeling distractors in training data as outliers of an optimization problem.
Our method successfully removes outliers from a scene and improves upon our baselines, on synthetic and real-world scenes.
Our technique is simple to incorporate in modern \NeRF frameworks, with 
few hyper-parameters.
It does not assume a priori knowledge of the types of distractors, and is instead focused on the optimization problem rather than pre-processing or modeling transient objects.
More results at  \url{https://robustnerf.github.io}.
\vspace*{-0.2cm}
\end{abstract}

\section{Introduction}
\label{sec:intro}
\vspace*{-0.1cm}

The ability to understand the structure of a \textit{static} 3D scene from 2D images alone is a fundamental problem is computer vision~\cite{tewari2022advances}.  It finds applications in AR/VR for mapping virtual environments~\cite{mobilenerf,zhu2022nice,rosinol2022nerf}, in autonomous robotics for action planning~\cite{adamkiewicz2022vision}, and in photogrammetry to create digital copies of real-world objects~\cite{reizenstein2021common}.

Neural fields~\cite{neuralfields} have recently revolutionized this classical task, by storing 3D representations within the weights of a neural network~\cite{sitzmann2019srns}.
These representations are optimized by back-propagating image differences.
When the fields store view-dependent \textit{radiance} and volumetric rendering is employed~\cite{neuralvolumes}, we can capture 3D scenes with photo-realistic accuracy, and we refer to the generated representation as Neural Radiance Fields, or NeRF~\cite{nerf}).

Training of NeRF models generally requires a large collection of images equipped with accurate camera calibration, which can often be recovered via structure-from-motion~\cite{schonberger2016structure}.
Behind its simplicity, NeRF hides several assumptions.
As models are typically trained to minimize error in RGB color space, it is of paramount importance that images are photometrically consistent -- two photos taken from the same vantage point should be \emph{identical} up to noise.
Unless one employs a method explicitly accounting for it~\cite{urf}, one should manually hold a camera's focus, exposure, white-balance, and ISO fixed.

\begin{figure}
\centering
\includegraphics[width=\linewidth]{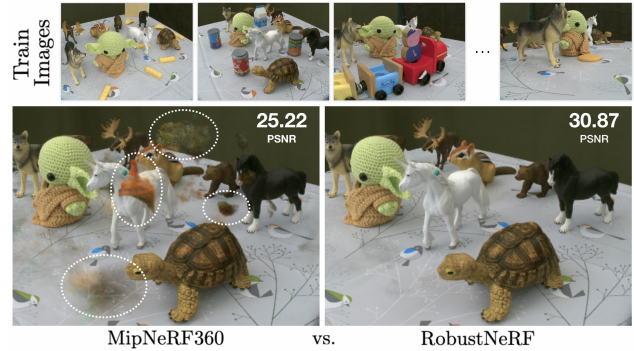}
\caption{
NeRF assumes photometric consistency in the observed images of  a scene.
Violations of this assumption, 
as with the images in the top row, yield reconstructed scenes with inconsistent content in the form of ``floaters'' (highlighted with ellipses).
We introduce a \textit{simple} technique that produces \textit{clean} reconstruction by automatically \textit{ignoring distractors} \textit{without explicit supervision}.
}
\vspace*{-0.30cm}
\label{fig:\currfilebase}
\end{figure}

However, properly configuring one's camera is not all that is required to capture high-quality NeRFs -- it is also important to avoid \textit{distractors}: anything that isn't persistent throughout the entire capture session.
Distractors come in many shapes and forms, from the hard-shadows cast by the operators as they explore the scene to a pet or child casually walking within the camera's field of view.
Distractors are tedious to \textit{remove} manually, as this would require pixel-by-pixel labeling.
They are also tedious to \textit{detect}, as typical NeRF scenes are trained from hundreds of input images, and the types of distractors are not known a priori.
If distractors are \textit{ignored}, the quality of the reconstruction scene suffers significantly; see \autoref{fig:teaser}.

In a typical capture session, it is difficult to 
to capture multiple images of the same scene from the same viewpoint, rendering distractors challenging to model mathematically.
As such, while view-dependent effects are what give NeRF their realistic look, \textit{how can the model tell the difference} between a distractor and a view-dependent effect?

Despite the challenges, the research community has devised several approaches to overcome this issue:
\begin{itemize}
\item If distractors are known to belong  to a specific class (e.g., people), one can remove them with a pre-trained semantic segmentation model~\cite{urf,tancik2022block} -- this process does \textit{not generalize} to ``unexpected'' distractors such as shadows.
\item One can model distractors as per-image \textit{transient} phenomena, and control the balance of transient/persistent modeling~\cite{nerfw} -- however, it is \textit{difficult to tune} the losses that control this Pareto-optimal objective.
\item One can model data in time (i.e., high-framerate video) and decompose
the scene into static and dynamic (i.e., distractor) components~\cite{ddnerf} -- but this clearly only applies to \textit{video} rather than photo collection captures.
\end{itemize}
\noindent
Conversely, we approach the problem of distractors by modeling them as \textit{outliers} in NeRF optimization. 

We analyze the aforementioned techniques through the lens of  robust estimation, 
allowing us to understand their behavior, and to  design a method that is not only simpler to implement but also more effective (see~\autoref{fig:teaser}).
As a result, we obtain a method that is straightforward to implement, requires minimal-to-no hyper-parameter tuning, and achieves state-of-the-art performance.
We evaluate our method: %
\begin{itemize}
\item quantitatively, in terms of reconstruction with synthetically, yet photo-realistically, rendered data;
\item qualitatively on publicly available datasets 
(often fine-tuned to work effectively with previous methods);
\item on a new collection of natural and synthetic scenes, including those autonomously acquired by a robot, allowing us to demonstrate the sensitivity of previous methods to hyper-parameter tuning.
\end{itemize}

\section{Related Work}
\vspace*{-0.1cm}
\label{sec:related}
We briefly review the basics and notation of Neural Radiance Fields.
We then  
describe recent progress in NeRF research, paying particular attention to techniques for modeling of static/dynamic scenes.

\paragraph{Neural Radiance Fields}
\label{sec:nerfreview}
A neural radiance field (NeRF) is a continuous volumetric representation of a 3D scene, stored within the parameters of a neural network $\params$.
The representation maps a position $\position$ and view direction $\dir$ to a \textit{view-dependent} RGB color and \textit{view-independent} density:
\begin{equation}
\begin{rcases}
\radiance(\position, \dir) \\
\density(\position)
\end{rcases} 
f(\position, \dir ; \params)
\label{eq:nerfmodel}
\end{equation}
This representation is trained from a collection,~$\{(\image_i, \pose_i)\}$, of images~$\image_i$ with corresponding calibration parameters~$\pose_i$~(camera extrinsics and intrinsics).

During training the calibration information is employed to convert each pixel of the image into a ray~$\ray {=} (\origin, \dir)$, and rays are drawn randomly from input images to form a training mini-batch~($\ray {\sim} \image_i$).
The parameters $\params$ are optimized to correctly predict the colors of the pixels in the batch via the L2 photometric-reconstruction loss:
\begin{align}
\loss{rgb}(\params) &= \sum_{i} \expect_{\ray \sim \image_i}
\left[
\loss{rgb}^{\ray, i}(\params) 
\right]
\\
\loss{rgb}^{\ray, i}(\params) &= ||\C(\ray; \params) - \image_i(\ray) ||_2^2 
\label{eq:nerfloss}
\end{align}
Parameterizing the ray as $\ray(t)=\origin + t\mathbf{d}$, the NeRF model image $\C(\ray; \params)$ is generated pixel-by-pixel volumetric rendering based on $\density(\cdot)$ and $\radiance(\cdot)$ (e.g., see \cite{volrendigest,nerf}).

\paragraph{Recent progress on NeRF models}
\label{sec:recentnerf}
NeRF models have recently been extended
in several ways.
A major thread has been the speedup of training~\cite{ingp, relufields} and inference~~\cite{snerg, mobilenerf}, enabling today's models to be trained in minutes~\cite{ingp}, and rendered on mobile in real-time~\cite{mobilenerf}.
While initially restricted to forward-facing scenes, researchers quickly found ways to model real-world $360^\circ$ scenes~\cite{mipnerf360, nerf++}, and to reduce the required number of images, via sensor fusion~\cite{urf} or hand-designed priors~\cite{regnerf}.
We can now deal with image artifacts such as motion blur~\cite{deblurnerf}, exposure~\cite{rawnerf}, and lens distortion~\cite{scnerf}.
And the requirement of (precise) camera calibrations is quickly being relaxed with the introduction of techniques for local camera refinement~\cite{barf, garf}, or  direct inference~\cite{relpose}.
While a NeRF typically represents geometry via volumetric density, there exist models custom-tailored to predict surfaces~\cite{neus, unisurf}, which can be extended to use predicted normals to significantly improve reconstruction quality~\cite{monosdf, neuris}.
Given high-quality normals~\cite{refnerf}, inferring the (rendering) structure of a scene becomes a possibility~\cite{samurai}.
We also note recent papers about additional applications to generalization~\cite{yu2021pixelnerf}, semantic understanding~\cite{nesf}, generative modeling~\cite{lolnerf}, robotics~\cite{adamkiewicz2022vision}, and text-to-3D~\cite{dreamfusion}.

\paragraph{Modeling non-static scenes}
\label{sec:staticvsdynamic}
\begin{figure*}[t]
\begin{center}
    \includegraphics[width=0.99\linewidth]{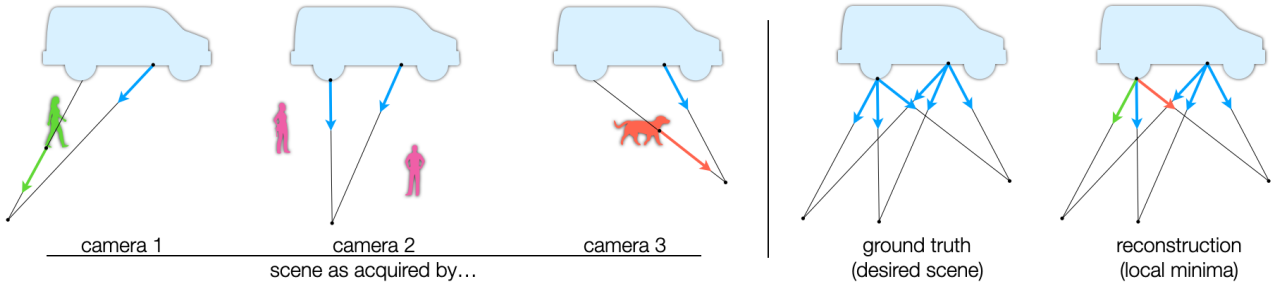}
\end{center}
\vspace{-1em}
\caption{
    \textbf{Ambiguity --}
    A simple 2D scene where a static object~(blue) is captured by three cameras.
    During the first and third capture the scene is not photo-consistent as a distractor was within the field of view.
    Not photo-consistent portions of the scene can end up being encoded as view-dependent effects -- even when we assume ground truth geometry.
}
\vspace*{-0.25cm}
\label{fig:\currfilebase}
\end{figure*}

For unstructured scenes like those considered here, the community has focused on reconstructing both static and non-static elements from video.
The most direct approach, treating time as an auxiliary input, leads to cloudy geometry and a lack of fine detail~\cite{xian2021space,gao2021dynamic}.
Directly optimizing per-frame latent codes as an auxiliary input has proved more effective~\cite{li2022neural,hypernerf,ddnerf}.
The most widely-adopted approach is to fit a time-conditioned deformation field mapping 3D points between pairs of frames~\cite{nsff,wang2021neural} or to a canonical coordinate frame~\cite{du2021neural,pumarola2021d,tretschk2021non,fang2022fast,liu2022devrf}.
Given how sparsely space-time is sampled, all methods require careful regularization, optimization, or additional training signals to achieve acceptable results.

Relatively little attention has been given to \emph{removing} non-static elements.
One common approach is to segment and ignore pixels which are likely to be distractors~\cite{tancik2022block, urf}.
While this eliminates larger objects, it fails to account for secondary effects like shadows.
Prior attempts to model distractors as outliers still leave residual cloudy geometry~\cite{nerfw}.

\section{Method}
\label{sec:method}
\vspace*{-0.1cm}

The classical NeRF training losses \eq{nerfloss} are effective for capturing scenes that are photometrically consistent, leading to the photo-realistic novel-view synthesis that we are now accustomed to seeing in recent research.
However, ``\textit{what happens when there are elements of the scene that are not persistent throughout the entire capture session?}''
Simple examples of such scenes include those in which an object is only present in some fraction of the observed images, or may not remain in the same position in all observed images.
For example, \autoref{fig:ambiguity}
depicts a 2D scene comprising a persistent object (the truck), along with several transient objects (e.g., people and a dog).
While rays in blue from the three cameras intersect the truck, the green and orange rays from cameras 1 and 3  intersect transient objects.
For video capture and spatio-temporal NeRF models, the persistent objects comprise the ``static'' portion of the scene, while the rest would be called the ``dynamic''.

\subsection{Sensitivity to outliers}
\label{sec:lambertian}
\vspace*{-0.1cm}

For Lambertian scenes, photo-consistent structure is view independent, as scene radiance only  depends on the incident light  \cite{SpaceCarving200}.
For such scenes, view-dependent NeRF models like~\eq{nerfmodel}, trained by minimizing \eq{nerfloss}, admit local optima in which transient objects are explained by view-dependent terms.
\autoref{fig:ambiguity} depicts this,
with the outgoing color corresponding to the memorized color of the outlier -- i.e. view-dependent radiance.
Such models exploit the view-dependent capacity of the model  to over-fit observations, effectively memorizing the transient objects.
One can alter the model to remove dependence on $\dir$, but the L2 loss remains problematic as least-squares (LS) estimators are sensitive to outliers, or heavy-tailed noise distributions.
 
Under more natural conditions, dropping the Lambertian assumption, the problem becomes more complex as {\em both} non-Lambertian reflectance phenomena and outliers can be explained as view-dependent radiance.
While we want the models to capture photo-consistent view-dependent radiance, 
outliers and other transient phenomena should ideally be ignored.
And in such cases, optimization with an L2 loss~\eq{nerfloss} yields significant errors in reconstruction; see~\autoref{fig:teaser}.
Problems like these are pervasive in NeRF model fitting, especially in  uncontrolled environments with complex reflectance, non-rigidity, or independently moving objects.

\subsection{Robustness to outliers}
\vspace*{-0.2cm}

\paragraph{Robustness via semantic segmentation}
One way to reduce outlier contamination during NeRF model optimization is to rely on an {\em oracle}~$\oracle$ that specifies whether a given pixel~$\ray$ from image~$i$ is an outlier,  and should therefore be excluded from the empirical loss, replacing \eq{nerfloss} with:
\begin{equation}
\loss{oracle}^{\ray, i}(\params) = 
\oracle_i(\ray)
\cdot
||\C(\ray; \params) - \image_i(\ray) ||_2^2 
\label{eq:oracleloss}
\end{equation}
In practice, a \textit{pre-trained} (semantic) segmentation network $\segmenter$ might serve as an oracle,  $\oracle_i {=} \segmenter(\image_i)$.
E.g., Nerf-in-the-wild~\cite{nerfw} employed a semantic segmenter to remove pixels occupied by people, as they are  outliers in the context of  photo-tourism.
Urban Radiance Fields~\cite{urf} segmented out sky pixels, while LOL-NeRF~\cite{lolnerf} ignored pixels not belonging to faces.
The obvious problem with this approach is the need for an oracle to detect arbitrary distractors.

\paragraph{Robust estimators}
Another way to reduce sensitivity to outliers is to replace the conventional L2 loss  \eq{nerfloss} with a \textit{robust loss} (e.g., \cite{regcourse,robustloss}), so that photometrically-inconsistent observations can be down-weighted during  optimization.
Given a robust kernel $\kernel(\cdot)$, we rewrite our training loss as:
\begin{equation}
\loss{robust}^{\ray, i}(\params) = 
\kernel(||\C(\ray; \params) - \image_i(\ray) ||_2) 
\label{eq:robustloss}
\end{equation}
where $\kernel(\cdot)$ is positive and monotonically increasing.
MipNeRF~\cite{mipnerf}, for example, employs an L1 loss $\kappa(\residual){=} | \residual |$, which provides some degree of robustness to outliers during NeRF training.
Given our analysis, a valid question is whether we can straightforwardly employ a robust kernel to approach our problem, and if so, given the large variety of robust kernels~\cite{robustloss}, which is the kernel of choice.

\begin{figure}[t!]
\vspace*{-0.1cm}
\centering
\includegraphics[width=\linewidth]{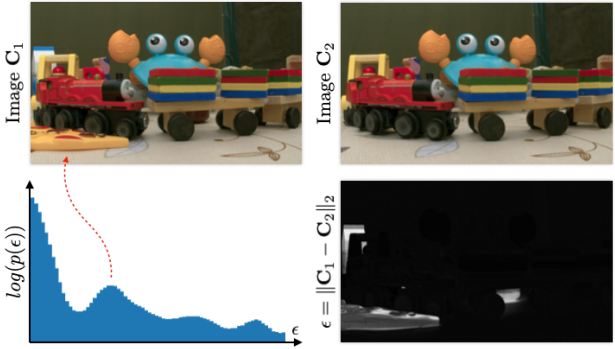}
\caption{\textbf{Histograms} -- 
Robust estimators perform well when the distribution of residuals agrees with the one implied by  the estimator (e.g., Gaussian for L2, Laplacian for L1).
Here we visualize the ground-truth distribution of residuals (bottom-left), which is hardly a good match with any simple parametric distribution.
}
\vspace*{-0.25cm}
\label{fig:histograms}
\end{figure}

Unfortunately, as discussed above, outliers and non-Lambertian effects can \textit{both} be modelled as view-dependent effects~(see \autoref{fig:histograms}).
As a consequence, with simple application of robust estimators it can be difficult to separate signal from noise.
\autoref{fig:kernels} shows examples in which outliers are removed, but fine-grained texture and  view-dependent details are also lost, or conversely, fine-grained details are preserved, but outliers cause artifacts in the reconstructed scene.
One can also observe mixtures of these cases in which details are not captured well, nor are outliers fully removed.
We find that this behaviour occurs consistently for many different robust estimators and parameter settings.

Training time can also be problematic. The robust estimator gradient w.r.t.\ model parameters can be expressed using the chain rule as
\begin{equation}
\left.
\frac{\partial \kernel(\residual(\params))}{\partial \params}
\right|_{\params^{(t)}}
=
\left.
\frac{\partial \kernel(\residual)}{\partial \residual}
\right|_{\residual(\params^{(t)})}
\cdot 
\left.
\frac{\partial \residual(\params)}{\partial \params}
\right|_{\params^{(t)}}
\label{eq:chainrule}
\end{equation}
The second factor is the classical NeRF gradient.
The first factor is the kernel gradient evaluated at the \textit{current} error residual $ \residual(\params^{(t)})$.
During training, large residuals can \textit{equivalently} come from high-frequency details that have not yet been learnt, or they may arise from outliers (see \autoref{fig:kernels}~(bottom)).
This explain why robust optimization, implemented as \eq{robustloss}, should not be expected to decouple high-frequency details from outliers.
Further, when \textit{strongly} robust kernels are employed, like redescending estimators, this also explains the loss of visual fidelity.  That is, because the gradient of (large) residuals get down-weighted by the (small) gradients of the kernel, \textit{slowing down} the learning of these fine-grained details (see \autoref{fig:kernels}~(top)).

\begin{figure}[t]
\begin{center}
\includegraphics[width=\linewidth]{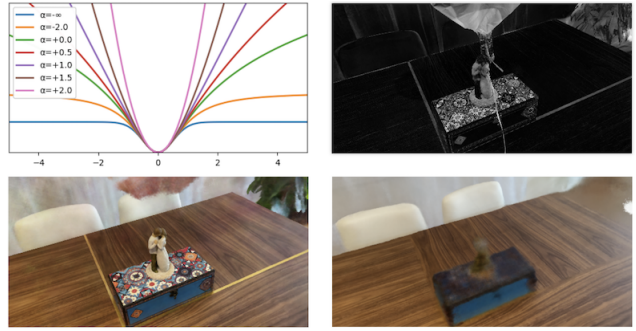}
\end{center}
\vspace*{-0.2cm}
\caption{
\textbf{Kernels --}
(top-left) Family of robust kernels~\cite{robustloss}, including L2~($\alpha{=}{2}$), Charbonnier~($\alpha{=}{1}$) and Geman-McClure~($\alpha{=}{-2}$). 
(top-right)
Mid-training, residual magnitudes are similar for distractors and fine-grained details, and pixels with large residuals are learned more slowly, as the gradient of re-descending kernels flattens out.
(bottom-right)
A too aggressive Geman-McClure in down-weighting large residuals 
removes both outliers and high-frequency detail.
(bottom-left)
A less aggressive Geman-McClure does not effectively remove outliers.}
\vspace*{-0.2cm}
\label{fig:\currfilebase}
\end{figure}

\subsection{Robustness via Trimmed Least Squares}
\vspace*{-0.1cm}

In what follows we advocate a form of iteratively reweighted least-squares (IRLS) with a Trimmed least squares (LS) loss for NeRF model fitting.

\vspace*{-0.05cm}
\paragraph{Iteratively Reweighted least Squares}
\begin{figure*}[h]
\centering
\includegraphics[width=0.925 \linewidth]{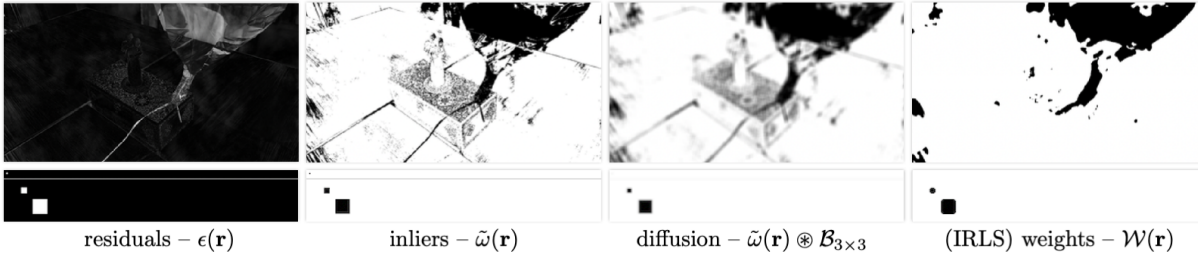}
\caption{\textbf{Algorithm} --
We visualize our weight function computed by residuals on two examples: 
(top) the residuals of a (mid-training) NeRF rendered from a \textit{training} viewpoint, 
(bottom) a toy residual image containing residual of small spatial extent (dot, line) and residuals of large spatial extent (squares).
Notice residuals with large magnitude but small spatial extent (texture of the box, dot, line) are included in the optimization, while weaker residuals with larger spatial extent are excluded.
Note that while we operate on patches, we visualize the weight function on the whole image to facilitate visualization.
}
\vspace*{-0.25cm}
\label{fig:\currfilebase}
\end{figure*}

IRLS is a widely used method for robust estimation that involves solving a sequence of weighted LS problems, the weights of which are adapted to reduce the influence of outliers.
To that end, at iteration $t$,  one can write the loss as
\begin{align}
\loss{robust}^{\ray, i}(\params^{(t)}) &= 
\weight(\residuals^{(t-1)}(\ray)) \cdot || \C(\ray; \params^{(t)}) - \image_i(\ray) ||_2^2
\label{eq:irls}
\nonumber
\\
\residuals^{(t-1)}(\ray) &= || \C(\ray; \params^{(t-1)}) - \image_i(\ray) ||_2
\end{align}
For weight functions given by 
$\weight(\residual) {=} \residual^{-1} \cdot \partial \kernel(\residual) / \partial \residual$
one can show that, under suitable conditions, the iteration converges to a local minima of \eq{robustloss} (see~\cite[Sec.~3]{regcourse}).

This framework admits a broad family of losses, including maximum likelihood estimators for heavy-tailed noise processes. Examples in \autoref{fig:kernels} include \label{fig:\currfilebase} the Charbonnier loss 
(smoothed L1), and more aggressive redescending estimators such as the Lorentzian or Geman-McClure \cite{robustloss}.
The objective in~\eq{oracleloss} can also be viewed as a weighted LS objective, the binary weights of which are provided by an oracle.
And, as discussed at length below, one can also view several recent methods like NeRFW~\cite{nerfw} and \ddnerf~\cite{ddnerf} through the lens of IRLS and weighted LS.

Nevertheless, choosing a suitable weight function $\weight(\residual)$ for NeRF optimization is non-trivial, due in large part to the intrinsic ambiguity between view-dependent radiance phenomena and outliers.
One might try to solve this problem by learning a neural weight function~\cite{acne}, although generating enough annotated training data might be prohibitive.
Instead, the approach taken below is to exploit inductive biases in the structure of outliers, combined with the simplicity of a robust, trimmed LS estimator.

\paragraph{Trimmed Robust Kernels}
Our goal is to develop a weight function for use in iteratively weighted LS optimization that is simple and captures useful inductive biases for NeRF optimization.
For simplicity we opt for a binary weight function with intuitive parameters that adapts naturally through model fitting so that fine-grained image details that are not outliers can be learned quickly.
It is also important to capture the structured nature of typical outliers, contrary to the typical i.i.d. assumption in most robust estimator formulations.
To this end, the weight function should capture spatial smoothness of the outlier process, recognizing that objects typically have continuous local support, and hence outliers are expected to occupy large, connected regions of an image~(e.g., the silhouette of a person to be segmented out from a photo-tourism dataset).

Surprisingly, a relatively simple weight function embodies these properties and performs extremely well in practice.
The weight function is based on so-called \textit{trimmed estimators} that are used in trimmed least-squares, like that used in trimmed ICP~\cite{trimmedicp}.
We first \textit{sort} residuals, and assume that residuals below a certain percentile are inliers.
Picking the 50\% percentile for convenience~(i.e., median), we define
\begin{equation}
\tilde{\weight}(\ray) = \residual(\ray) \leq \threshold_\residual ~,
~~~
\threshold_\residual = \text{Median}_\ray \{\residual(\ray)\} ~.
\label{eq:trimming}
\end{equation}

To capture spatial smoothness of outliers we spatially diffuse inlier/outlier labels $\weight$ with a $3{\times}3$ box kernel $\mathcal{B}_{3\!\times\!3}$.
Formally, we define
\begin{align}
\mathcal{W}(\ray) = \tilde\weight(\ray) | (\tilde\weight(\ray) \circledast \mathcal{B}_{3\times3}) \geq \threshold_\circledast ~,
~~~ \threshold_\circledast = 0.5 ~.
\label{eq:smoothing}
\end{align}
This helps to avoid classifying high-frequency details as outliers, allowing them to be captured by the NeRF model during optimization (see~\autoref{fig:algorithm}).

While the trimmed weight function (\ref{eq:smoothing}) improves the robustness of model fitting, it sometimes misclassifies fine-grained image details early in training where the NeRF model first captures coarse-grained structure. These localized texture elements may emerge but only after very long training times.
We find that stronger inductive bias to spatially coherence allows fine-grained details to be learned more quickly.
To that end, we  aggregate the detection of outliers on $16\!{\times}\! 16$ neighborhoods;
i.e., we label entire $8\!\times\! 8$ patches as outliers or inliers based on the behavior of $\mathcal{W}$ in the $16 {\times} 16$ neighborhood of the patch .
Denoting the $N {\times} N$ neighborhood of pixels around $\ray$ as $\mathcal{R}_{N}(\ray)$, we define
\newcommand{\sray}{\mathbf{s}}
\begin{align}
\weight(\mathcal{R}_8(\ray)) &= \mathcal{W}(\ray) | \expect_{\sray \sim \mathcal{R}_{16}(\ray)}
\left[ 
\mathcal{W}(\sray)
\right] \geq \threshold_\mathcal{R} \, , ~~~ \threshold_\mathcal{R}=0.6 \, .
\label{eq:patching}
\end{align}
This robust weight function 
evolves during optimization, as one expects with IRLS where the weights are a function of the residuals at the previous iteration.
That is, the labeling of pixels as inliers/outliers \textit{changes} during training, and settles around masks similar to the one an oracle would provide as training converges (see~\autoref{fig:residuals}).

\begin{figure}[t!]
\centering
\includegraphics[width=\linewidth]{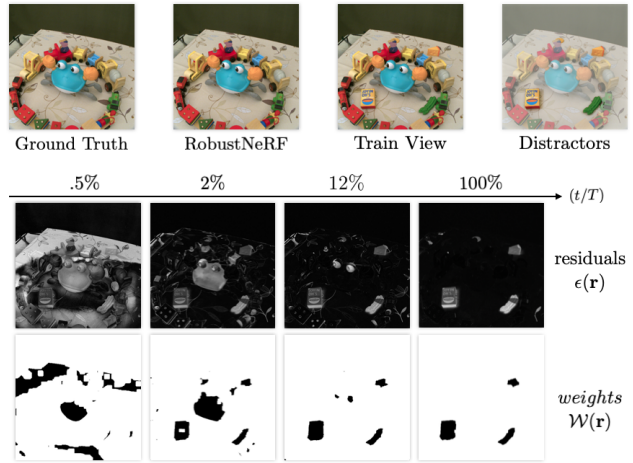}
\caption{
\textbf{Residuals -- }
For the dataset shown in the top row, we visualize the dynamics of the \RobustNeRF training residuals, which show how over time the estimated distractor weights go from being random~($(t/T){=}0.5\%$) to identify distractor pixels~($(t/T){=}100\%$) without any explicit supervision.
}
\vspace*{-0.25cm}
\label{fig:\currfilebase}
\end{figure}

\section{Experiments}
\label{sec:experiments}
\vspace*{-0.1cm}

We implement our robust loss function in the MultiNeRF codebase~\cite{multinerf2022} and apply it to \mipNeRFthreesixty~\cite{mipnerf360}.
We dub this method ``\RobustNeRF''.
To evaluate \RobustNeRF, we compare  against baselines on several scenes containing different types of distractors.
Where possible, we quantitatively compare reconstructions to held-out, distraction-free images;
we report three metrics, averaged across held-out frames, namely, 
PSNR, SSIM~\cite{ssim2004}, and LPIPS~\cite{zhang2018perceptual}.

We compare different methods on two collections of scenes, i.e., those provided by the authors of \DDNeRF, and \textit{novel} datasets described below.
We also present a series of illustrative experiments on synthetic scenes, shedding light on \RobustNeRF's efficacy and inner workings.

\subsection{Baselines}
\vspace*{-0.1cm}

We compare \RobustNeRF to variants of \mipNeRFthreesixty optimized with different loss functions~($L_2$, $L_1$, and Charbonnier).
These variants serve as natural baselines for models 
with limited or no robustness to outliers.We also compare to \DDNeRF, a recent method for reconstructing dynamic scenes from monocular \textit{video}. Unlike our method, \DDNeRF is designed to \textit{reconstruct} distractors rather than discard them.
While \DDNeRF is presented as a method for monocular video, it does not presuppose  temporal continuity,  and can be directly applied to unordered images.
We omit additional comparisons to \NeRFW as its performance falls short of~\DDNeRF~\cite{ddnerf}.
For more details on model training, see the \SupplementaryMaterial{~(\autoref{sec:trainingdetails})}.

\subsection{Datasets -- \autoref{fig:dataset}}
\vspace*{-0.1cm}

In addition to scenes from \DDNeRF, we introduce a set of natural and synthetic scenes. They facilitate the evaluation of \RobustNeRF's effectiveness on illustrative use cases, and they enable empirical analysis under controlled conditions.

\begin{figure}
\centering
\includegraphics[width=\linewidth]{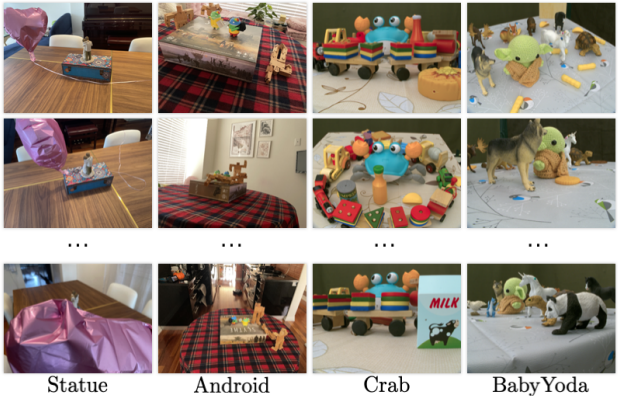}
\caption{\textbf{Dataset -- }
Sample training images showing the distractors in each scene. 
Statue and Android were acquired manually, and the others with a robotic arm. 
In the robotic setting we have pixel-perfect alignment of distractor vs.\ distractor-free images.
}
\vspace*{-0.25cm}
\label{fig:\currfilebase}
\end{figure}

\newcommand{\lpips}{\scalebox{0.8}{LPIPS$\downarrow$}}
\newcommand{\mssim}{\scalebox{0.8}{MS-SSIM$\uparrow$}}
\newcommand{\ssim}{\scalebox{0.8}{SSIM$\uparrow$}}
\newcommand{\psnr}{\scalebox{0.8}{PSNR$\uparrow$}}
\newcommand{\traintime}{\scalebox{0.8}{Train~Time$\downarrow$}}

\begin{figure*}[t!]
\centering
\resizebox{\linewidth}{!}{ %
\begin{tabular}{@{}l|ccc|ccc|ccc|ccc@{}}
& \multicolumn{3}{c|}{\color{brown}\bf Statue} 
    & \multicolumn{3}{c|}{\color{dark_green} \bf Android} 
    & \multicolumn{3}{c|}{\color{red}\bf Crab} 
    & \multicolumn{3}{c}{\color{blue}\bf BabyYoda} 
\\
                             & \lpips & \ssim & \psnr & \lpips & \ssim & \psnr  & \lpips & \ssim & \psnr  & \lpips & \ssim & \psnr  \\
\midrule
\mipNeRFthreesixty($L_2$)    & 0.36  & 0.66 & 19.09 &  0.40  & 0.65 & 19.35 &  0.27 & 0.77  & 25.73  &  0.31  &  0.75 &  22.97      \\
\mipNeRFthreesixty($L_1$)    & 0.30  & 0.72 & 19.55 &  0.40  & 0.66 & 19.38 &  0.22 & 0.79  & 26.69  &  0.22   & 0.80  &  26.15      \\
\mipNeRFthreesixty(Ch.)      & 0.30  & 0.73 & 19.64 &  0.40 & 0.66 & 19.53 &  0.21  & 0.80  & 27.72  &  0.23 &  0.80  &  25.22     \\
\ddnerf                      & 0.48  & 0.49 &19.09 &   0.43 & 0.57 & 20.61  & 0.42  & 0.68 &  21.18  &  0.44 &  0.65  & 17.32    \\
\textbf{\RobustNeRF}         & \textbf{0.28 } & \textbf{0.75} & \textbf{20.89} &  \textbf{0.31} & \textbf{0.65} & \textbf{21.72} &  \textbf{0.21}  &   \textbf{0.81} & \textbf{30.75} & \textbf{0.20}  & \textbf{0.83}   & \textbf{30.87}  \\
\midrule
\mipNeRFthreesixty(clean)   & 0.19  &0.80 & 23.57 &  0.31 &  0.71 & 23.10 &   0.16      &  0.84    &   32.55    &  0.16     &   0.84   &   32.63    \\
\end{tabular}
}
\\[.5em]
\includegraphics[width=\linewidth]{fig/\currfilebase}
\vspace{-1em}
\captionof{figure}{
\textbf{Evaluation on Natural Scenes --}
\RobustNeRF 
outperforms baselines and  
\ddnerf~\cite{ddnerf} on novel view synthesis with  real-world captures.
The table provides a quantitative comparison of
\RobustNeRF, \ddnerf and \mipNeRFthreesixty using different reconstruction losses.
The last row reports \mipNeRFthreesixty trained on a distractor-free version of each dataset, giving an upperbound for \RobustNeRF performance. 
We also visualize samples from each scene rendered with each of the methods. See Supplementary Material for more samples.}
\label{fig:\currfilebase}
\vspace{-0.25cm}
\end{figure*}

\vspace{-0.05cm}
\paragraph{Natural scenes}
We capture seven natural scenes exemplifying different types of distractors.
Scenes are captured in three settings, on the street, in an apartment and in a robotics lab.
Distractor objects are moved, or are allowed to move, between frames to simulate capture over extended periods of time.
We vary the number of unique distractors  from 1 (Statue) to 150 (BabyYoda),
and their  movements.
Unlike prior work on monocular video, frames are captured without a clear temporal ordering (see \autoref{fig:dataset}). 
The other three (i.e., Street1, Street2, and Gloss)  include view-dependence effects, the results of which are shown in the supplementary material.
We also capture additional frames \textit{without distractors} to enable quantitative evaluations.
Camera poses are estimated using COLMAP~\cite{schoenberger2016mvs}.
A full description of each scene in the \SupplementaryMaterial{~(\autoref{app:sec:naturalscenes})}.

\vspace{-0.1cm}
\paragraph{Synthetic scenes}
To further evaluate \RobustNeRF, we generate synthetic scenes using the Kubric dataset generator~\cite{greff2021kubric}.
Each scene is constructed by placing a set of simple geometries in an empty, texture-less room.
In each scene, a subset of objects remain fixed while the other objects~(i.e., distractors) change position from frame to frame.
By varying the number of objects, their size, and the way they move, we control the level of distraction in each scene.
We use these scenes to examine \RobustNeRF's sensitivity to its hyperparameters, see~\SupplementaryMaterial{~(\autoref{app:sec:sensitivity_to_hyperparameters})}.

\subsection{Evaluation}
\vspace*{-0.1cm}

\begin{figure*}[ht]
\centering
\setlength\tabcolsep{1.8pt}
\resizebox{\linewidth}{!}{
\begin{tabular}{l|ccc|ccc|ccc|ccc|ccc}
\multicolumn{1}{c}{}  & \multicolumn{3}{c|}{Car}  & \multicolumn{3}{c|}{Cars} & \multicolumn{3}{c|}{\bf{Bag}} & \multicolumn{3}{c|}{Chairs} & \multicolumn{3}{c}{Pillow}
\\ & \lpips & \mssim & \psnr& \lpips & \mssim & \psnr& \lpips & \mssim & \psnr& \lpips & \mssim & \psnr& \lpips & \mssim & \psnr \\
\midrule
NeRF-W \cite{nerfw} &  .218 &  .814 & 24.23 &  .243 &  .873 & 24.51 &  .139 &  .791 & 20.65 &  .150 &  .681 & 23.77 &  .088 &  .935 & 28.24 \\
NSFF \cite{nsff} & .200 & .806 & 24.90 & .620 & .376 & 10.29 & .108 & .892 & 25.62 & .682 & .284 & 12.82 & .782 & .343 & 4.55 \\
NeuralDiff \cite{neuraldiff}  &  .065 &  .952 & 31.89 &  .098 &  .921 & 25.93 &  .117 &  .910 & 29.02 &  .112 &  .722 & 24.42 &  .565 &  .652 & 20.09 \\
\ddnerf~\cite{ddnerf} & {.062} & {.975} & {34.27} & {.090} & {.953} & {26.27} & {.076} & {.979} & {34.14} & {.095} &  .707 & {24.63} & {.076} & {.979} & {36.58}       \\
\textbf{\RobustNeRF} & \bf{.013} & \bf{.988} & \bf{37.73} & \bf{.063} & \bf{.957} & \bf{26.31} & \bf{.006} & \bf{.995} & \bf{41.82} & \bf{.007} & \bf{.992} & \bf{41.23} & \bf{.018} & \bf{.990} & \bf{38.95} \\ 
\end{tabular}
} %
\\[.25em]
\includegraphics[width=\linewidth]{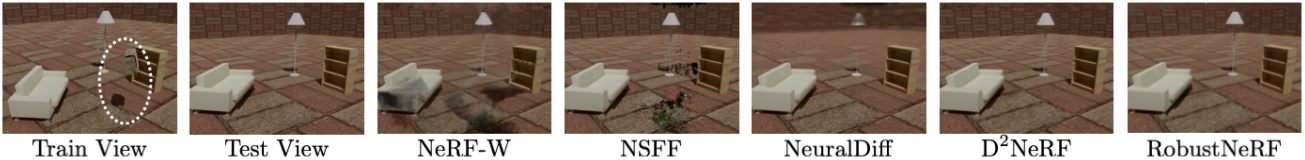}
\vspace{-.94em}
\captionof{figure}{
\textbf{Evaluations on \ddnerf Synthetic Scenes --}
Quantitative and qualitative evaluations on the Kubric synthetic dataset introduced by \ddnerf, consisting of 200 training frames (with distractor) and 100 novel views for evaluation (without distractor).
}
\vspace*{-0.275cm}
\label{fig:\currfilebase}
\end{figure*}

We evaluate \RobustNeRF on its ability to \textit{ignore} distractors while accurately reconstructing the static elements of a scene.
We train \RobustNeRF, \DDNeRF, and variants of \mipNeRFthreesixty on scenes where distraction-free frames are available.
Models are \textit{trained} on frames with distractors and \textit{evaluated} on distractor-free frames.

\vspace{-0.1cm}
\paragraph{Comparison to \mipNeRFthreesixty{} -- \autoref{fig:sota}}
On natural scenes, \RobustNeRF generally outperforms variants of \mipNeRFthreesixty by 1.3 to 4.7 dB in PSNR.
As $L_2$, $L_1$, and Charbonnier losses weigh all pixels equally, the model is forced to represent, rather than ignore, distractors as ``clouds'' with view-dependent appearance.
We find clouds to be most apparent when distractors remain stationary for multiple frames.
In contrast, \RobustNeRF's loss isolates distractor pixels and assigns them a weight of zero (see \autoref{fig:residuals}).
To establish an upper bound on reconstruction accuracy, we train \mipNeRFthreesixty with Charbonnier loss on distraction-free versions of each scene,
the images for which are taken from (approximately) the same viewpoints.
Reassuringly, \RobustNeRF when trained on distraction-free frames, achieves nearly identical accuracy; see \autoref{fig:limitations}.

While \RobustNeRF consistently outperforms \mipNeRFthreesixty, the gap is smaller in the Apartment scenes (Statue, Android) than the Robotics Lab scenes (Crab, BabyYoda).
This can be explained by challenging background geometry, errors in camera parameter estimation, and imperceptible changes to scene appearance.
For further discussion, see the \SupplementaryMaterial{~(\autoref{app:sec:comparison_mipnerf360})}.

\vspace{-0.05cm}
\paragraph{Comparison to \ddnerf~-- \autoref{fig:sotaddnerf}}
Quantitatively, \RobustNeRF matches or outperforms \DDNeRF by as much as 12 dB PSNR depending on the number of unique outlier objects in the capture. Results on
\ddnerf real scenes are provided in the supplementary material for qualitative comparison.
In Statue and Android, 1 and 3 non-rigid objects are moved around the scene, respectively.
\DDNeRF is able to model these objects and thus separate them from the scenes' static content.
In the remaining scenes, a much larger pool of 100 to 150 unique, non-static objects are used -- too many for \DDNeRF to model effectively.
As a result, ``cloud'' artifacts appear in its static representation, similar to those produced by \mipNeRFthreesixty.
In contrast, \RobustNeRF identifies non-static content as outliers and omits it during reconstruction.
Although both methods use a similar number of parameters, \DDNeRF's peak memory usage is 2.3x higher than \RobustNeRF and 37x higher when normalizing for batch size.
This is a direct consequence of model architecture: \DDNeRF is tailored to simultaneously modeling static and dynamic content and thus merits higher complexity.
To remain comparable, we limit image resolution to 0.2 megapixels for all experiments. 

\begin{figure}[t]
\centering
\resizebox{\linewidth}{!}{ %
\begin{tabular}{@{}lcccc@{}}
& \lpips  & \ssim & \psnr & Updates to PSNR=30
\\ \midrule
\mipNeRFthreesixty ($L_2$) &  0.31 &  0.75    &  22.97   &   --\\
+ robust~\eq{trimming}  &  0.39 & 0.60   &  18.21    &     --\\
+ smoothing~\eq{smoothing} &   0.22 &  0.81   & 30.01    &   250K \\
+ patching~\eq{patching} & \textbf{0.21} & \textbf{0.81}   &  \textbf{30.75}    &   70K  \\
\midrule
oracle (clean) & 0.16 & 0.84 & 32.55 & 25K \\
\end{tabular}
} %
\\
\includegraphics[width=\linewidth]{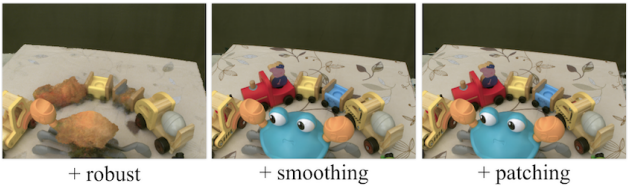}
\captionof{figure}{\textbf{Ablations --}
Blindly trimming the loss causes details to be lost.
Smoothing recovers fine-grained detail, while patch-based evaluation speeds up training and adds more detail.
Patching enables the model to reach PSNR of $30$, almost $4\times$ faster.
}
\vspace*{-0.25cm}
\label{fig:\currfilebase}
\end{figure}

\vspace{-1em}
\paragraph{Ablations -- \autoref{fig:ablations}}
We ablate elements of the \RobustNeRF loss on the crab scene, comparing to an upper bound on the reconstruction accuracy of  \mipNeRFthreesixty trained on distractor-free (clean) images from identical viewpoints.
Our trimmed estimator
(\ref{eq:trimming}) 
successfully eliminates distractors at the expense of high frequency texture and a lower PSNR.
With smoothing
(\ref{eq:smoothing}), 
fine details are recovered, at the cost of longer training times.
With the spatial window
(\ref{eq:patching}),
\RobustNeRF training time is on-par with \mipNeRFthreesixty.
We also ablate patch size and the trimming threshold (see Supplementary Material); we find that \RobustNeRF is
insensitive to trimming threshold, and that
reducing the patch size offsets the gains from smoothing and patching.

\paragraph{Sensitivity -- \autoref{fig:limitations}}
We find that \RobustNeRF is remarkably robust to the amount of clutter in a dataset.
We define an image as ``cluttered'' if it contains some number of distractor  pixels.
The figure shows how  the reconstruction accuracy of \RobustNeRF and \mipNeRFthreesixty depends on the fraction of training images with distractors, keeping the training set size constant.
As the fraction increases, \mipNeRFthreesixty's accuracy steadily drops from 33 to 25 dB, while \RobustNeRF's remains steadily above 31 dB throughout.
In the distraction-free regime, we find that \RobustNeRF mildly under-performs \mipNeRFthreesixty, both in reconstruction quality and the time needed for training.
This follows from the statistical inefficiency induced by the trimmed estimator~\eq{trimming}, for which a percentage of pixels will be discarded even if they do not correspond to distractors.

\color{black}

\vspace*{-0.1cm}
\section{Conclusions}
\label{sec:conclusions}
\vspace*{-0.1cm}

We address a central problem in training NeRF models, namely, optimization in the presence of distractors, such as  transient or moving objects and photometric phenomena that are not persistent throughout the capture session. 

\begin{figure}
\vspace*{-0.185cm}
\centering
\includegraphics[width=\linewidth]{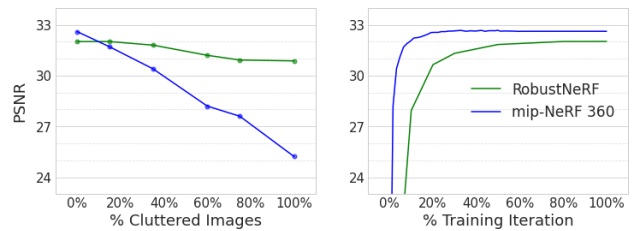}
\caption{
\textbf{Sensitivity and Limitations -- }
(left) Reconstruction accuracy for BabyYoda as we increase the fraction of train images with  distractors.
(right) Accuracy vs training time on \textit{clean} BabyYoda images~(distractor-free).
}
\vspace*{-0.255cm}
\label{fig:\currfilebase}
\end{figure}

Viewed through the lens of robust estimation, 
we formulate training as a form of iteratively re-weighted least squares, with  
a variant of trimmed LS, and an inductive bias on the smoothness of the outlier process.
\RobustNeRF is surprisingly simple, yet effective on a wide range of datasets.
\RobustNeRF is shown to outperform recent state-of-the-art methods~\cite{mipnerf360, ddnerf}, qualitatively and quantitatively, on a suite of synthetic datasets, common benchmark datasets, and new datasets captured by a robot, allowing fine-grained control over distractors for comparison with previous methods.
While our experiments explore robust estimation in the context of  \mipNeRFthreesixty, the \RobustNeRF loss can be incorporated within other NeRF models.

\vspace*{-0.11cm}
\paragraph{Limitations} 
While \RobustNeRF performs well on scenes with distractors, the loss entails some statistical inefficiency.
On clean data, this yields somewhat poorer reconstructions, often taking longer to train~(see \autoref{fig:limitations}).
Future work will consider very small distractors,
which may require adaptation of the spatial support used for outlier/inlier decisions.
It would also be interesting to learn a neural weight function, further improving \RobustNeRF;  active learning may be useful in this context.
Finally, it would be interesting to include our robust loss in other NeRF frameworks.

\vspace*{0.085cm}

\noindent
{\bf Acknowledgements}
We thank Pete Florence and Konstantinos Rematas for helpful feedback, and Tianhao Wu for help with \ddnerf experiments.

{
    \small
    \bibliographystyle{ieee_fullname}
    \bibliography{macros,main}
}
\clearpage
\twocolumn[\centering\Large\textbf{\thetitle \\ (Supplementary Material)}\vspace{1.0em}]

\setcounter{section}{6}

\subsection{Dataset Description}
\label{app:sec:datasets}

To investigate \RobustNeRF and its baselines, we capture and generate a collection of natural and synthetic scenes.
With the goal of reconstructing the \emph{static} elements of a scene, we capture frames both with and without distractors present.
We describe the details of the capture below.

\subsubsection{Natural Scenes}
\label{app:sec:naturalscenes}

We introduce four natural scenes, two captured in an apartment setting, and two in a robotics lab. See \autoref{tab:dataset_details} for key details.

\paragraph{Apartment (Statue \& Android)}
\label{app:sec:apartmentscenes}
To mimic a casual home scenario, we capture two tabletop scenes in an apartment using a commodity smartphone.
Both captures focus on one or more objects on a table top, with photos taken from different viewpoints from a hemisphere of directions around the objects of interest.
A subset of objects on the table move from photo to photo as described below.
The photos within each scene do not have a clear temporal order.

The capture setup is as follows.
We employ an iPhone 12 mini and use ProCamera v15 to control camera exposure settings.
We use a fixed shutter speed of 1/60, 0.0 exposure bias, and a fixed ISO of 80 or 200 for the Statue and Android scenes, respectively.
We use the iPhone's standard wide lens with an aperture of f/1.6 and resolution of 4032x3024.
A tripod is used to reduce the effects of the rolling shutter.

The Android dataset comprises 122 cluttered photos and 10 clean photos
(i.e., with no distractors).
This scene depicts two Android robot figures standing on a board game box, which in turn is sitting on a table with a patterned table cloth.
We pose three small wooden robots atop the table in various ways in each cluttered photo to serve as distractors.

For the Statue scene, we capture 255 cluttered photos and 19 clean photos.
The scene depicts a small statue on top of a highly-detailed decorative box on a wooden kitchen table.
To simulate a somewhat persistent distractor, we float a balloon over the table which, throughout the capture, naturally changes its position slightly with each photo.
Unlike the Android scene, where distractors move to entirely new poses in each frame, the balloon frequently inhabits the same volume of space for multiple photos.
The decorative box and kitchen table both exhibit fine grained texture details. 

\begin{figure}
\centering

\resizebox{\linewidth}{!}{ %
\begin{tabular}{@{}l|cccccc@{}}
           & \# Clut. & \# Clean & \# Extra & Paired? & Res.      & Setting      \\
\midrule
Android     & 122      & 122        & 10       & No      & 4032x3024 & Apartment    \\
Statue    & 255      & 132        & 19       & No      & 4032x3024 & Apartment    \\
Crab       & 109      & 109      & 194      & Yes     & 3456x3456 & Robotics Lab \\
BabyYoda   & 109      & 109      & 202      & Yes     & 3456x3456 & Robotics Lab \\
\bottomrule
\end{tabular}
}

\caption{
    \textbf{Natural Scenes} --
    Key facts about natural scenes introduced in this work.
    Includes number of paired photos with (\# Clut.) and without (\# Clean) distractors.
    Extra photos (\# Extra) do not contain distractors and are taken from unpaired camera poses.
}
\label{tab:\currfilebase}
\end{figure}

We run COLMAP's~\cite{schonberger2016structure} Structure-from-Motion pipeline using the \texttt{SIMPLE\_RADIAL} camera model.
While COLMAP's camera parameter estimates are only approximate, we find that they are sufficient for training \NeRF models with remarkable detail.

The apartment scenes are considerably more challenging to reconstruct than the robotics lab scenes (described below).
An accurate \NeRF reconstruction must model not only the static, foreground content but also the scene's background.
Unlike the foreground, each object in the background is partially over- or underexposed and appears in a limited number of photos.
We further found it challenging to maintain a controlled, static scene during capture.
As a result, some objects in the background move by a small, unintended amount between photos (e.g., see \autoref{fig:residual_android}).

\paragraph{Robotics Lab (Crab \& BabyYoda)}
\label{app:sec:roboticslabscenes}
In an effort to control confounding factors in data acquisition, we capture two scenes in a Robotics Lab setting.
In these scenes, we employ a robotic arm to randomly position a camera within 1/4 of the hemisphere over a table.
The table is placed in a closed booth with constant, indoor lighting.
A series of toys are placed on the table, a subset of which are glued to the table's surface to prevent them from moving.
Between photos, distractor toys on the table are removed and/or new distractor toys are introduced.

For capture, we use a Blackfly S GigE camera with a TECHSPEC 8.5mm C Series fixed length lens.
Photos are center-cropped from their original resolution of 5472x3648 to 3456x3456 to eliminate lens distortion.
We capture 12-bit raw photos with an aperture of f/8 and exposure time of 650 ms.
Raw photos are automatically color-calibrated afterwards according to a reference color palette.

In each scene, we capture 109 pairs of photos from identical camera poses, one with distractors present and another without.
This results in a large number of unique distractors which are challenging to model directly.
This further allows us to investigate the counterfactual: What if distractors were \emph{not} present?
We further capture an additional $\sim$200 photos from random viewpoints, not aligned with those for training and without distractors, for the purposes of evaluation.
In total, because the placement of objects is done manually, one capture session often takes several hours.

\subsubsection{Synthetic Scenes}
\label{app:sec:syntheticscenes}
\begin{figure}[t]
\centering
\includegraphics[width=\linewidth]{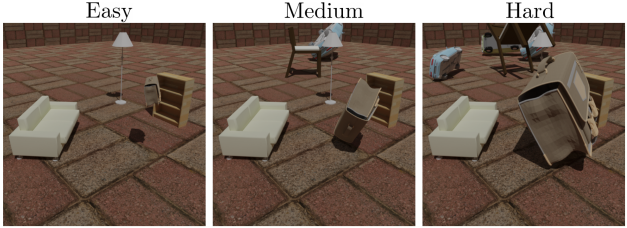}
\caption{
    \textbf{Synthetic Kubric Scenes} --
    Example Kubric synthetic images for three datasets with different ratio of outlier pixels. The sofa, lamp, and bookcase are static objects in all three setups. The easy setup has 1 small distractor, the medium setup has 3 medium distractors, and the hard setup has 6 large distractors.
}
\label{fig:\currfilebase}
\end{figure}

We generate three Kubric \cite{greff2021kubric} scenes similar to the \ddnerf synthetic scenes with different difficulty levels: easy, medium, and hard. These datasets are used to ablate our method with control on the proportion of outlier occupancy (see Sec.\  \ref{app:sec:sensitivity_to_hyperparameters}). 

 Each dataset contains 200 cluttered images for training  and 100 clean images for evaluation. In all three scenes the static objects include a sofa, a lamp and a bookshelf. \autoref{fig:synth} shows one example image from the training set for each dataset. The easy scene contains only one small distractor object (a bag). This dataset is similar to Kubric Bag dataset of \ddnerf. The medium scene has three distractors (a bag, a chair, and a car) which are larger in size and hence the outlier occupancy is  $4\times$ the outlier occupancy of the easy scene. The hard scene has six large distractors (a bag, a chair, and four cars). They occupy on average $10\times$ more pixels than the easy setup, covering  roughly half of each image. 

\subsection{Training Details}
\label{sec:trainingdetails}

While camera parameters are estimated on the full-resolution imagery, we downsample images by 8x for each natural scene dataset.
While \mipNeRFthreesixty and \RobustNeRF are capable of training on high resolution photos, we limit the resolution to accommodate \DDNeRF.
Unless otherwise stated, we train on all available cluttered images, and evaluate on a holdout set; i.e., 10 images for Android; 19 for the Statue dataset; 194 for Crab; and 202 for the BabyYoda dataset
(see \autoref{tab:dataset_details}).

\paragraph{\RobustNeRF}
\label{app:sec:robustnerf}
We implement \RobustNeRF by incorporating our proposed loss function into the MultiNeRF  codebase~\cite{multinerf2022}, replacing  \mipNeRFthreesixty's~\cite{mipnerf360} reconstruction loss.
All other terms in the loss function, such as regularizers, are included as originally published in \mipNeRFthreesixty.

We train \RobustNeRF for 250,000 steps with the Adam optimizer, using a batch size of 64 image patches randomly sampled from training images.
Each pixel within a 16x16 patch contributes to the loss function, except those identified as outliers
(see \autoref{fig:residuals} for a visualization).
The learning rate is exponentially decayed from 0.002 to 0.00002 over the course of training with a warmup period of 512 steps.

Our model architecture comprises a proposal Multilayer Perceptron (MLP) with 4 hidden layers and 256 units per layer, and a \NeRF MLP with 8 hidden layers, each with 1024 units.
We assign each training image a 4-dimensional GLO vector to account for unintended appearance variation.
Unless otherwise stated, we use the robust loss hyperparameters given in the main body of the paper.
All models are trained on 16 TPUv3 chips over the course of 9 hours.

\paragraph{\mipNeRFthreesixty~\cite{mipnerf360}}
\label{app:sec:mipnerf360}
We use the reference implementation of \mipNeRFthreesixty from the MultiNeRF codebase.
Similar to \RobustNeRF, we train each variant of  \mipNeRFthreesixty with the Adam optimizer, using the same number of steps, batch size, and learning rate schedule.
\mipNeRFthreesixty uses a random sample of 16384 rays per minibatch.
Proposal and \NeRF MLP depth and width are identical to those for \RobustNeRF.
Training hardware and duration are also the same as \RobustNeRF.

\paragraph{\ddnerf~\cite{ddnerf}}
\label{app:sec:ddnerf}
We use the reference implementation of \DDNeRF~\cite{ddnerf} provided by the authors.
Model architecture, hierarchical volume sampling density, and learning rate are the same as published in~\cite{ddnerf}.
As in the original work, we train the model for 100,000 iterations with a batch size of 1024 rays, though over the course of 3 hours. Due to hardware availability, we employ four NVIDIA V100 GPUs in place of the A100 GPUs used in the original work.

Images are kept in the order of provided by the file system (i.e., ordered by position information alphanumerically). However, this image order is not guaranteed to represent a continuous path in space since the images were not captured along a continuous path, but rather at random locations. Below we discuss the effects of random ordering versus ordering the views along a heuristically identified path.

\begin{figure*}[t!]
\centering
\includegraphics[width=\textwidth]{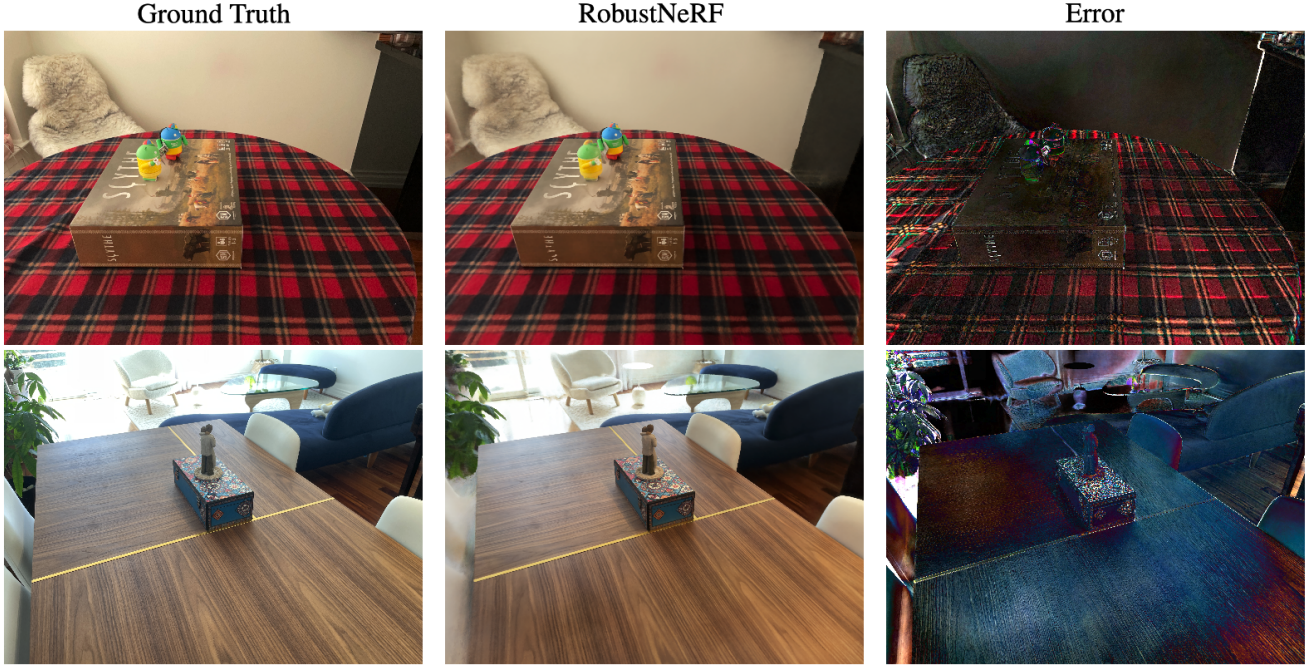}
\caption{
    \textbf{Challenges in Apartment Scenes -- }
  Each row, from left to right, shows a ground truth photo, a \RobustNeRF render, and the difference between the two. Best viewed in PDF.
   (Top) Note the fold in the table cloth in ground truth image and the lack of fine-grained 
    detail on the covered chair in the background.
    The table cloth moved during capture, and the background was not captured thoroughly enough for a high-fidelity reconstruction.
    (Bottom) The ground truth image for the Statue dataset exhibits overexposure and color calibration issues, and hence do not exactly match the \RobustNeRF render.
}
\vspace*{-0.1cm}
\label{fig:\currfilebase}
\end{figure*}

\DDNeRF training is controlled by five key hyperparameters, namely, skewness ($k)$, which encourages a binarization loss to favor static explanations, and four regularization weights that scale the skewed binarization loss ($\lambda_s$), ray regularization loss ($\lambda_r$), static regularization loss ($\lambda_{\sigma^s}$), and the view-correlated shadow field loss ($\lambda_\rho$). A hyperparameter search is performed in \ddnerf for 16 real world scenes to identify combinations best suited for each scene, and four primary configurations of these parameters are identified as optimal. In particular, the first configuration (i.e., $k = 1.75$, $\lambda_s = 1e^{-4} \rightarrow 1e^{-2}$, $\lambda_r = 1e^{-3}$, $\lambda_{\sigma^s} = 0$, and $\lambda_\rho = 1e^{-1}$) was reported to be most effective across the largest number of scenes real world (10 of 16). We additionally conduct a tuning experiment (see \autoref{fig:ddnerf_tune}) and confirm the first configuration as best suited. We apply this configuration in all additional \ddnerf experiments.

\subsection{Experiments}
\label{app:sec:experiments}

\subsubsection{Comparison to \mipNeRFthreesixty}
\label{app:sec:comparison_mipnerf360}
In experiments on natural scenes, as reported in \autoref{fig:sota}, the performance gap between \mipNeRFthreesixty (Ch.) and \RobustNeRF is markedly higher for the two scenes captured in the robotics lab (i.e., Crab, BabyYoda), compared to those in the apartment (i.e., Statue, Android).
We attribute this to the difficulty in reconstructing the apartment scenes, regardless of the presence of distractors.
This statement is supported by metrics for reconstruction quality of a \mipNeRFthreesixty model trained on clean, distractor-free photos.
In particular, while \mipNeRFthreesixty achieves over 32 dB PSNR on Crab and BabyYoda scenes, its PSNR is nearly 10 dB lower on Statue and Android.

Upon closer inspection of the photos and our reconstructions, we identified several reasons for this.
First, the apartment scenes contain non-trivial background content with 3D structure.
As the background was not the focus of these captures, background content is poorly reconstructed by all models considered.
Second, background content illuminated by sunlight is overexposed in some test images (see \autoref{fig:residual_android}).
While this challenge has already been addressed by RawNeRF~\cite{rawnerf}, we do not address it here as it is not a focus of this work.
Lastly, we find that some static objects were unintentionally moved during our capture.
The most challenging form of this is the movement of a table cloth prominently featured in the Android scene which lead to perturbed camera parameter estimates (e.g., see \autoref{fig:residual_android}).

\subsubsection{Comparison to \DDNeRF}
\label{app:sec:comparison_ddnerf}
\begin{figure}
\centering

\resizebox{\linewidth}{!}{ %

\begin{tabular}{@{}l|ccc|ccc@{}}
    & \multicolumn{3}{c|}{\color{red}\bf Crab} 
    & \multicolumn{3}{c}{\color{blue}\bf BabyYoda} 
\\
                             & \lpips & \ssim & \psnr  & \lpips & \ssim & \psnr  \\
\midrule
Order 1                      &  0.43 & 0.66  & 20.19  &  0.44  &  0.66  &  18.17      \\
Order 2                    &  0.42 & 0.68  & 20.95  & 0.44    &  0.66  &  17.13      \\
\midrule
\end{tabular}
}

\includegraphics[width=\linewidth]{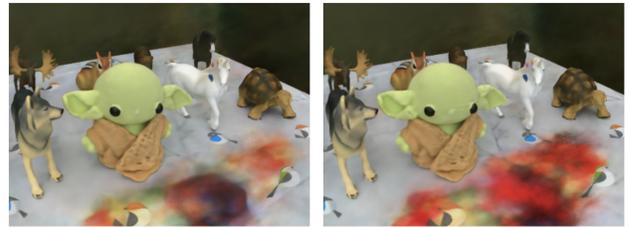}
\caption{
\textbf{Effect of Image Order on \DDNeRF} --
As this model is based on space-time NeRFs~\cite{hypernerf}, to make it compatible with our setting we create a 'temporal' indexing of the photos. 
Here, we visualize:
(left) with our heuristic ordering;
(right) with another random order.  We observe similar distractor-related artifacts in both cases.
}
\vspace{-.5cm}
\label{fig:\currfilebase}
\end{figure}

\begin{figure}
\centering

\resizebox{\linewidth}{!}{ %
\begin{tabular}{@{}l|ccc|ccc@{}}
    & \multicolumn{3}{c|}{\color{brown}\bf Statue} 
    & \multicolumn{3}{c}{\color{red}\bf Crab} 
\\
                    & \lpips & \ssim & \psnr  & \lpips & \ssim & \psnr  \\
\midrule
Config 1            & 0.48  & 0.49  & 19.09  &  0.42  &  0.68  &  21.18      \\
Config 2            & 0.49  & 0.48  & 18.20  & 0.51  & 0.59  & 17.02 \\
Config 3            & 0.51  & 0.47  & 18.28  & 0.46  & 0.63  & 19.01 \\
Config 4            & 0.49  & 0.48  & 18.18  & 0.49  & 0.58  & 16.77 \\
\midrule
\end{tabular}
} \\ [1ex]

\begin{tabular}{c | c c c c c} 
\hline
Config $\#$ & $k$ & $\lambda_s$ & $\lambda_r$ & $\lambda_{\sigma^s}$  & $\lambda_\rho$ \\ [0.5ex] 
\hline
Config 1 & 1.75 & 1e-4 $\rightarrow$ 1e-2 & 1e-3 & 0 & 1e-1 \\ 
Config 2 & 3 & 1e-4 $\Rightarrow$ 1 & 1e-3 & 0 & 1e-1 \\
Config 3 & 2.75 & 1e-5 $\Rightarrow$ 1 & 1e-3 & 0 & - \\
Config 4 & 2.875 & 5e-4 $\Rightarrow$ 1 & 0 & 0 & - \\ 
\hline
\end{tabular}

\caption{\textbf{\ddnerf HParam Tuning} --
The performance of \ddnerf{} is heavily influenced by the choice of hyperparameters. In particular, optimal choices of hyperparameters are noted to be strongly influenced by the amount of object and camera motion, as well as video length. We tune by applying four recommended configurations, and identify the first as optimal across the Statue and Crab datasets. Please note that $\rightarrow$ indicates linear increase in value and $\Rightarrow$ indicates exponential increase in value.
}
\label{fig:\currfilebase}
\end{figure}

Unlike \RobustNeRF, \ddnerf makes use of a time signal in the form of provided appearance and warp IDs to generate a code as additional input to the HyperNeRF model.
This explicitly models dynamic content alongside the static component of the scene. Two assumptions of \ddnerf are broken in our datasets: 1) the objects sporadically appear (by design); and 2) the views are not necessarily captured in a video-like order. Sporadic object appearance is central to our task, so we do not ablate this property. 
However, we do evaluate the effect of heuristically reordering camera views according to z position and radial angle of the robotic arm, thereby producing an image order for an imagined "continuous" path.
As a control, we pseudorandomly scramble the view order, and train \ddnerf in both settings. The results for BabyYoda and Crab can be seen in \autoref{fig:ddnerf}. We observe no consistent discernable improvement in performance as a result of view reordering and hypothesize that the major hurdle for \ddnerf is rather the modeling of sporadic artifacts. 

We also evaluate the effect of applying the four hyperparameter configurations provided by \ddnerf \cite{ddnerf}. 
We observe, as expected, that the first configuration performs best across our datasets.
Due to limited access to appropriate compute architecture for \ddnerf, we were not able to tune every scene, but selected configuration 1 for all experiments as it performed best in 10/16 real world scenes for \ddnerf as well as tuning experiments on two of our example datasets as see in in \autoref{fig:ddnerf_tune}.

\subsubsection{Sensitivity to Hyperparameters}
\label{app:sec:sensitivity_to_hyperparameters}
\begin{figure}
\centering
\includegraphics[width=0.75\linewidth]{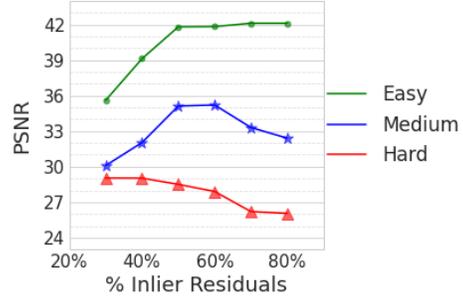}
\caption{
    \textbf{Sensitivity to $\threshold_\residual$ --}
    \RobustNeRF's reconstruction quality as a function of $\threshold_\residual$ on scenes with different inlier/outlier proportions.
    Overestimating $\threshold_{\residual}$ increases training time without affecting final reconstruction accuracy.
}
\label{fig:\currfilebase}
\end{figure}

We find that the choice of thresholds and filter sizes, described in Section~\ref{sec:method}, suffices for a wide range of datasets.
As long as the threshold $\threshold_\residual$ is greater than the proportion of outlier pixels in a dataset, \RobustNeRF will reliably identify and ignore outlier pixels; see \autoref{fig:sensitivity_to_outliers}. Easy has less than $10\%$ outlier pixels so any  $\threshold_\residual$ less than $80\%$ works. In the medium case at least a $\threshold_\residual$ of $60\%$ is required to remove  the outliers. In the hard case $44\%$ of pixels are on average occupied so any  $\threshold_\residual$ above $50\%$ has worse results. Training with less than $50\%$ of the loss slows down training significantly. Therefore, we observe that after the 250k iterations the model has not converged yet. On average training with $30\%$ of loss requires twice the number of training iterations to catch up.
In contrast, \DDNeRF requires careful, manual hyperparameter tuning for each scene (e.g., see \autoref{fig:ddnerf_tune}) for several hyperparameters.
In our experiments, we found that a single setting of neighborhood and patch sizes works well across all scenes.
We present model performance as a function of both hyperparameters on Crab in \autoref{tab:rebuttal_ablations}.
Larger neighborhood sizes are better regularizers, and we are bounded by the amount of device memory available.

\begin{figure}[t]
\small
\centering
\begin{tabular}{@{}lccccc@{}}
    Neigh./Patch & 4/2 & 8/4   & 16/2  & 16/4  & 16/8  \\
    \midrule
    $\threshold_\mathcal{R}=0.6$       & 18.3  & 23.08 & 30.22 & 30.35 & 30.75 \\
    $\threshold_\mathcal{R}=0.8$       & 28.28  & 30.7    & 30.72    & 30.69    & 30.72    \\
\end{tabular}
\\
\captionof{figure}{
    \textbf{Sensitivity to hyper-parameters.}
    PSNR on distractor-free frames on the Crab dataset as a function of \RobustNeRF's neighborhood size, patch size, and $\threshold_\mathcal{R}$.
}
\vspace*{-0.4cm}
\label{tab:\currfilebase}
\end{figure}

\subsubsection{View-dependent effects}
\label{app:sec:view_dependent}

We experimentally observed that \RobustNeRF performs similarly to \mipNeRFthreesixty in reconstructing scenes with non-Lambertian materials, semi-transparent objects, and soft shadows.
These phenomena are present in the Statue scene (tabletop is \textit{glossy}), and the toys in the Crab and BabyYoda scenes which cast soft \emph{shadows}.

To further emphasize these qualities, we  include results for additional scenes with glass, metallic, and reflective objects in  \autoref{fig:rebuttal_viewdep}. The first scene is captured with our Robotic rig, similar to Crab and BabyYoda scenes.  It contains a mirror, a shiny cylinder, a transparent vase and a glossy ceramic mug. The other two datasets are captured in the wild. One is with a mirror while pedestrians are moving (as distractions). The last scene  contains a transparent pitcher as the object of interest, while the photographer's body parts appear in the photos as the distractors.

\begin{figure}[t]
\centering

\includegraphics[width=\linewidth]{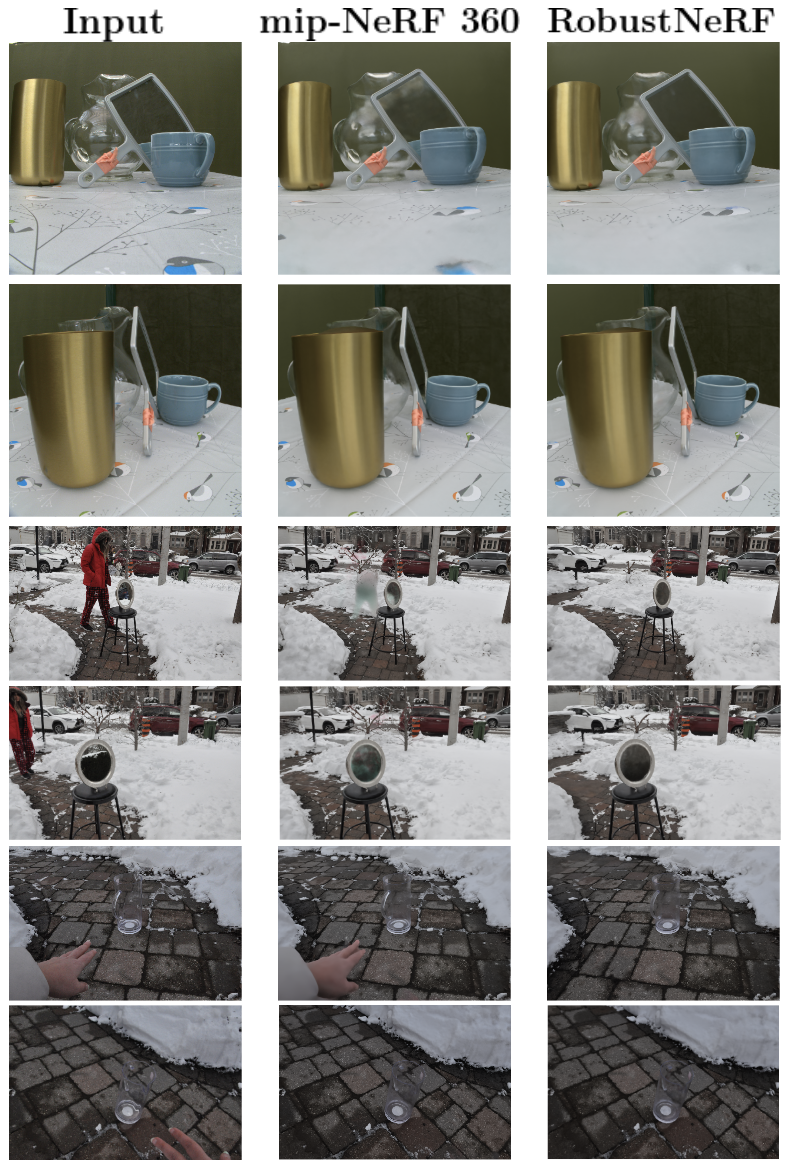}
\captionof{figure}{
    \textbf{Qualitative results on scenes with view-dependent effects.}
    \RobustNeRF naturally captures view-dependent effects in scenes with (3rd-6th rows) and without (1st and 2nd row) distractors.
}
\vspace*{-0.2cm}
\label{fig:\currfilebase}
\end{figure}

\subsubsection{More Qualitative Results}
\label{app:sec:sota_more}

We render images from different NerF models from more  viewpoints from each of our datasets to further expand the comparison with baselines, \ddnerf, and \RobustNeRF. Looking at Figures  \ref{fig:bigsota_statue} through
\ref{fig:bigsota_yoda} one can see that \ddnerf is only able to remove the outliers when there is a single distractor object (Statue dataset) and it fails on the other three datasets. The Android dataset has three wooden robots with articulated joints as distractors, and even in this setup where the texture of the distractor objects are similar to one another, \ddnerf fails to fully remove the outliers. In comparison, \RobustNeRF is able to remove the outliers irrespective of their number and diversity. 

For all four datasets \mipNeRFthreesixty, with either L1, L2, or Charbonnier loss, fails to detect the outliers;
one can see 'clouds' or even distinct floaters for these methods. The worst performing loss is L2, as expected. L1 and Charbonnier behave similarly in terms of outlier removal. Changing the loss to \RobustNeRF eliminates the  floaters and artifacts in all datasets. Video renderings for these scenes are also included in the zipfile with the supplementary material. The floaters in \mipNeRFthreesixty are  easier to resolve in the videos.

We have also experimented on the \ddnerf natural scenes in ~\cite{ddnerf}.
The qualitative samples are shown in \autoref{fig:rebuttal_d2nerf}.
We find that \RobustNeRF produces plausible, distraction-free models. 

\begin{figure}[t]
\centering

\includegraphics[width=\linewidth]{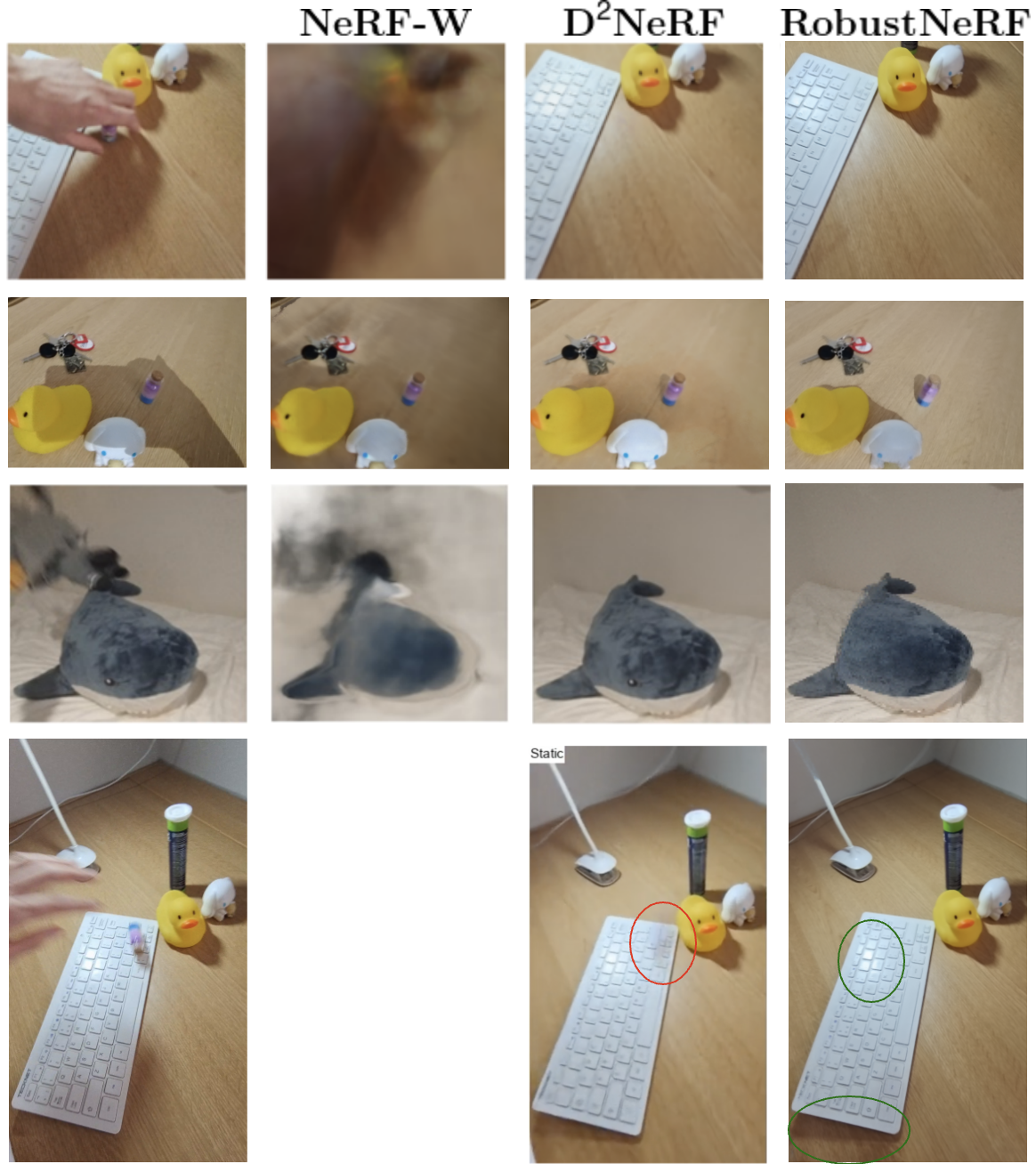}
\captionof{figure}{
\textbf{Qualitative results on \ddnerf Pick scene.}
Renders of static model components. 
Results for \NeRFW and \ddnerf are provided by \cite{ddnerf}. %
Note how \RobustNeRF naturally captures specular reflections and shadows (green, right).
}
\vspace*{-0.2cm}
\label{fig:\currfilebase}
\end{figure}

\clearpage
\begin{figure*}[t!]
\centering
\includegraphics[width=\textwidth]{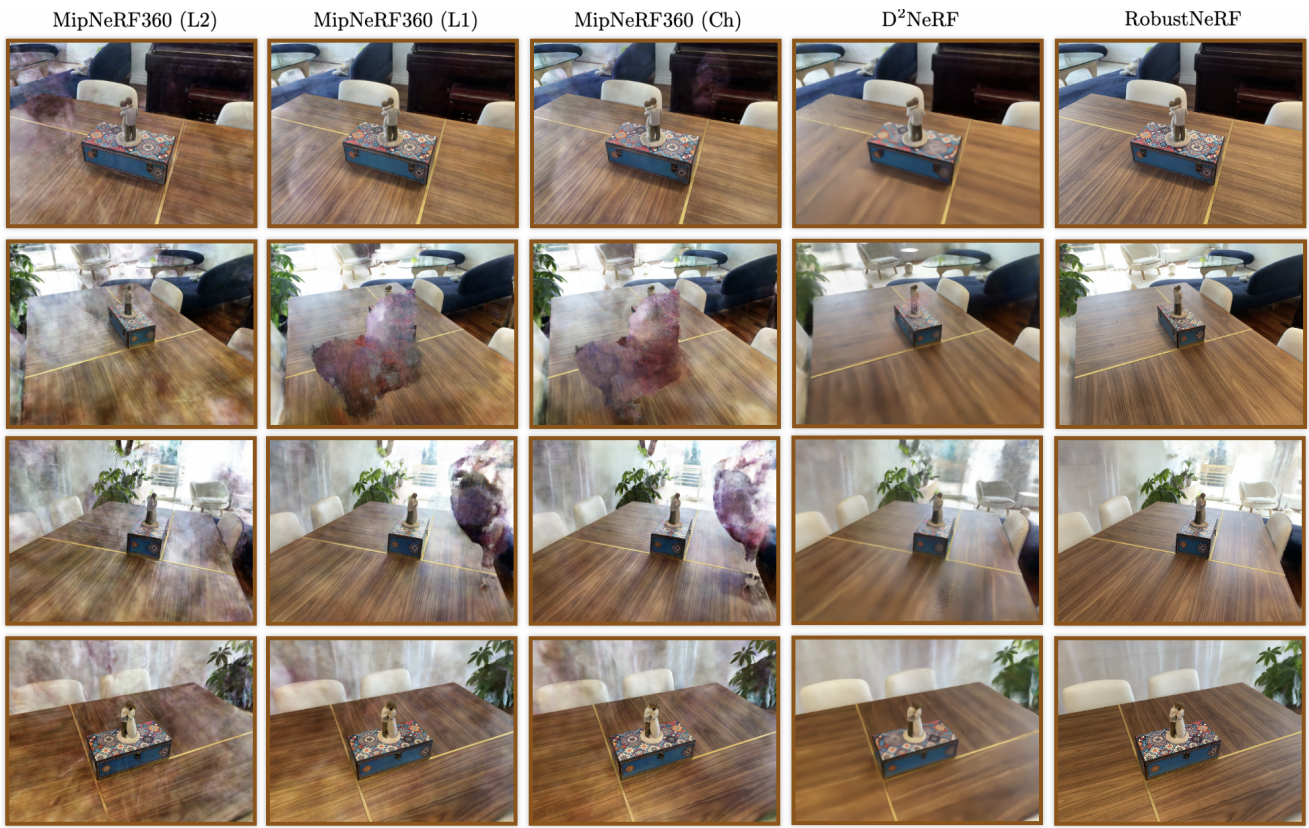}
\caption{
    \textbf{Statue --}
    Qualitative results on Statue. 
    It is helpful to zoom in to see details.
}
\label{fig:\currfilebase}
\end{figure*}

\begin{figure*}[b!]
\centering
\includegraphics[width=\textwidth]{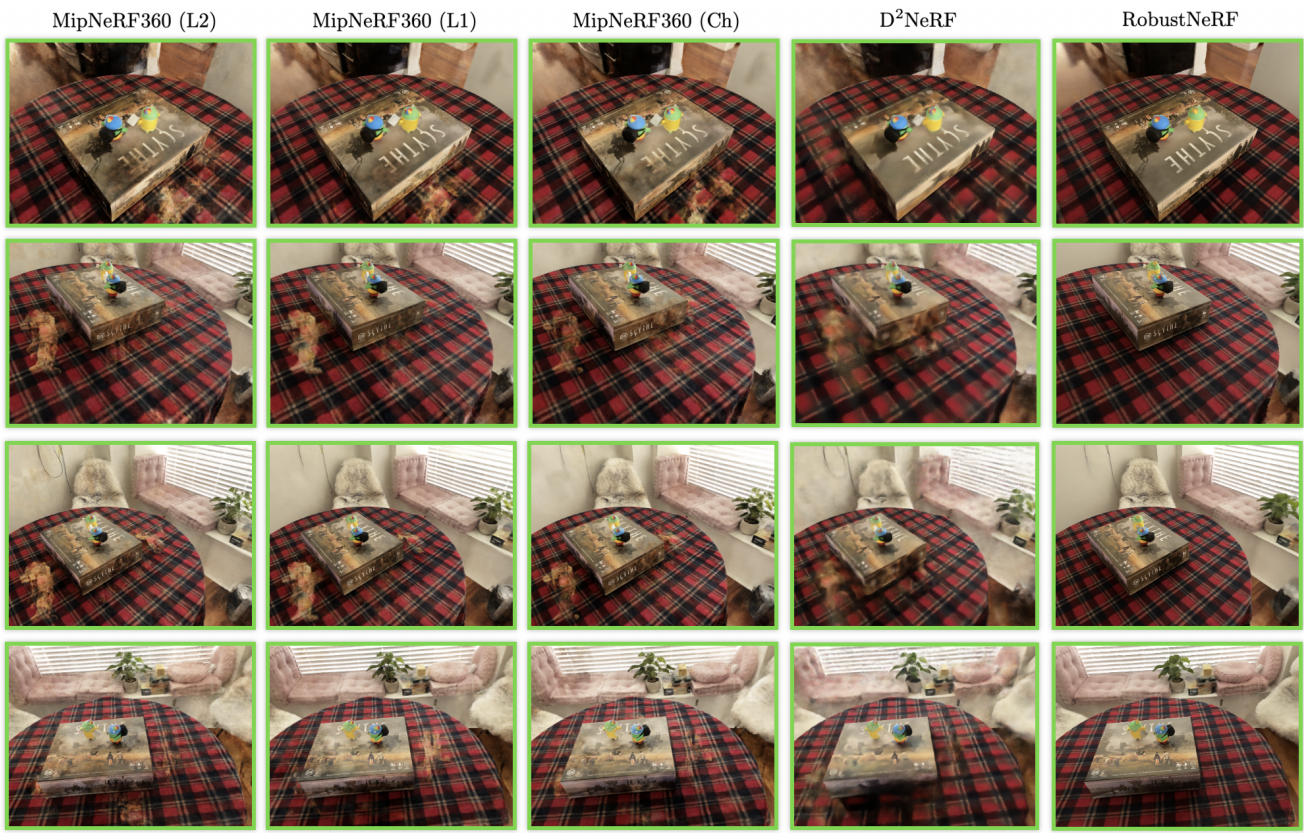}
\caption{
    \textbf{Android --}
    Qualitative results on Android.
    It is helpful to zoom in to see details.
}
\label{fig:\currfilebase}
\end{figure*}

\begin{figure*}[t]
\centering
\includegraphics[width=\textwidth]{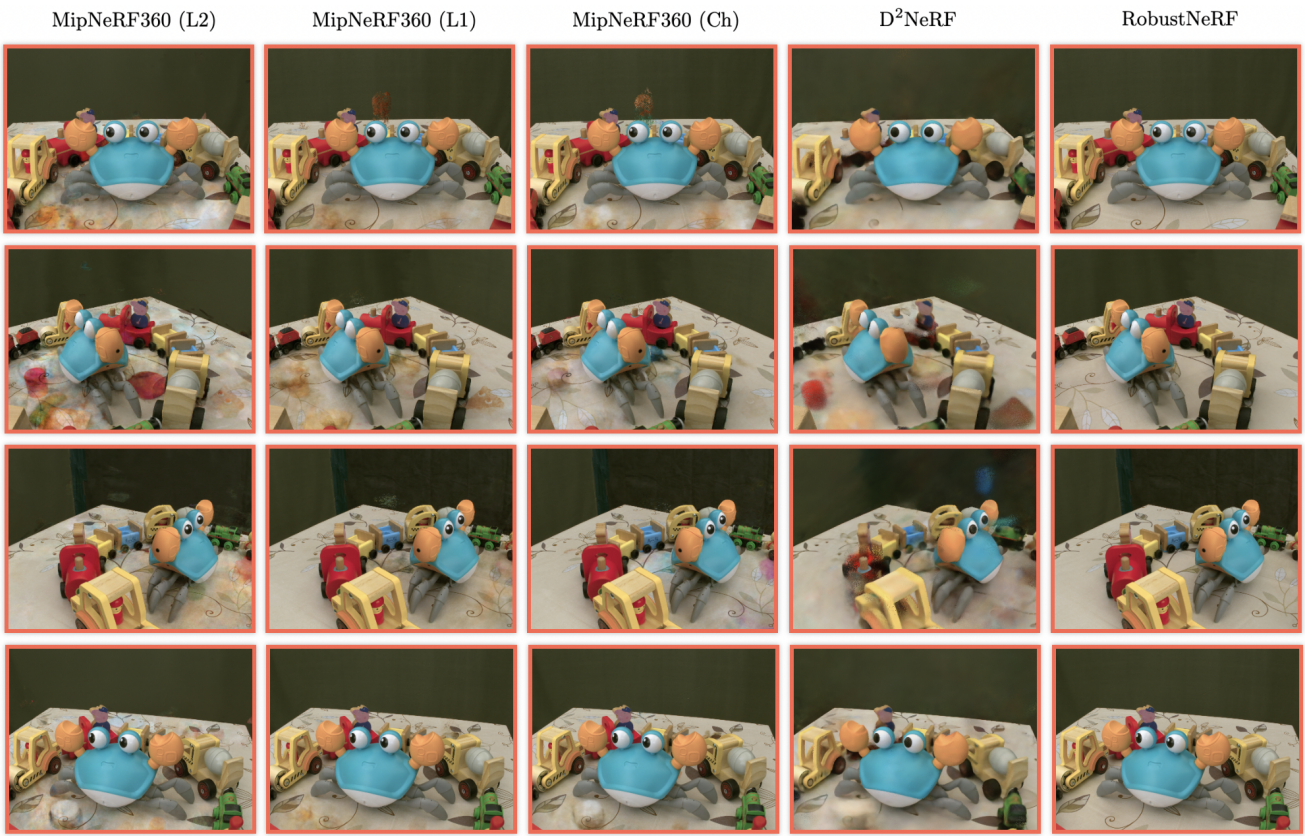}
\caption{
    \textbf{Crab --}
    Qualitative results on Crab. 
    It is helpful to zoom in to see details.
}
\label{fig:\currfilebase}
\end{figure*}

\begin{figure*}[b]
\centering
\includegraphics[width=\textwidth]{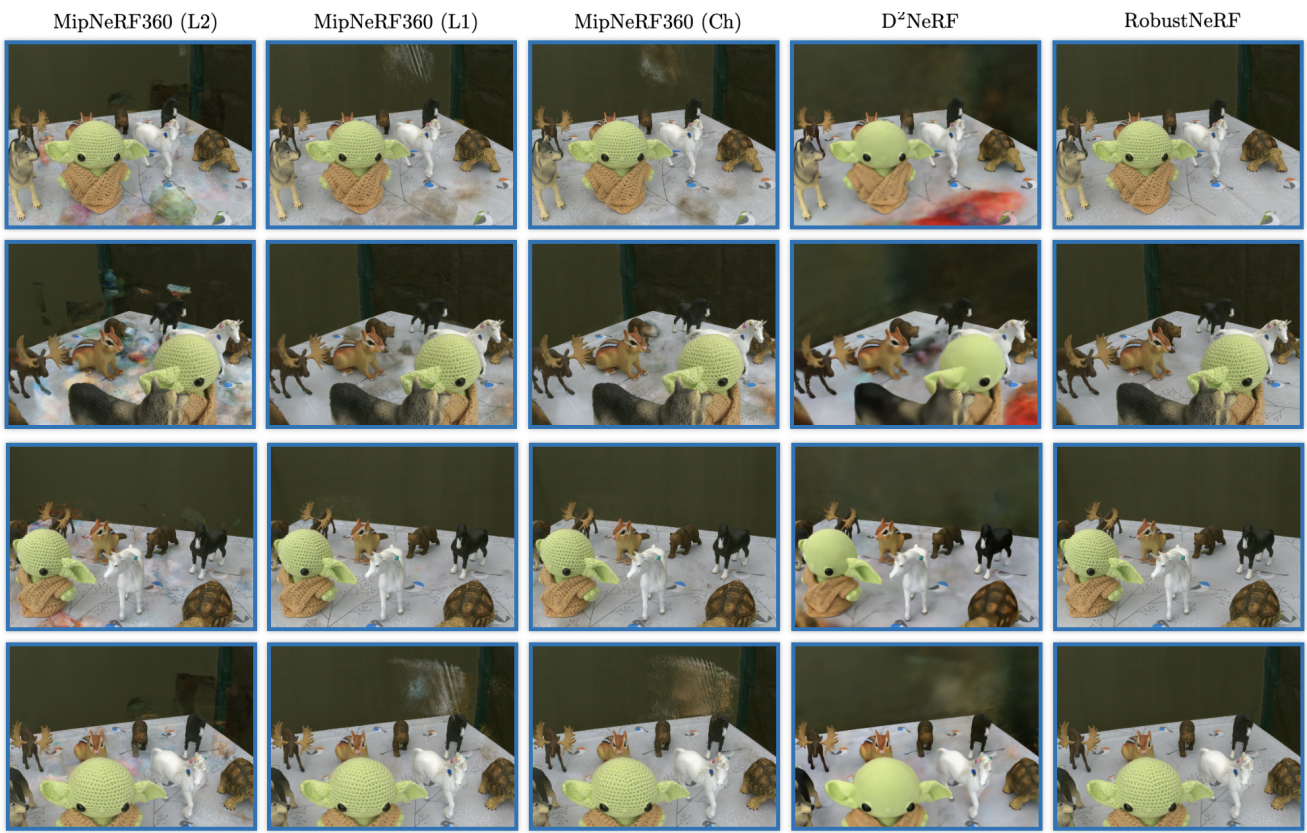}
\caption{
    \textbf{BabyYoda --}
    Qualitative results on BabyYoda. 
    It is helpful to zoom in to see details.
}
\label{fig:\currfilebase}
\end{figure*}

\end{document}